\documentclass[runningheads]{llncs}
\usepackage{graphicx}
\usepackage[toc]{appendix}
\usepackage{enumitem}
\usepackage{multirow}
\usepackage[T1]{fontenc}
\usepackage{lmodern}
\usepackage{textcomp, gensymb}
\usepackage{cite}
\usepackage[margin=3.75cm]{geometry}
\usepackage[linesnumbered]{algorithm2e}
\usepackage{hyperref}
\usepackage{comment}
\usepackage{silence}
\begin{document}
\title{Semantic Enrichment of the Quantum Cascade Laser Properties in Text- A  Knowledge Graph Generation Approach}
\titlerunning{Knowledge Graph Generation for QCL Properties from Text}
%
%
\author{Deperias Kerre\inst{1,3}\orcidID{0000-0002-7437-6735} \and
Anne Laurent\inst{1}\orcidID{0000-0003-3708-6429} \and
Kenneth Maussang\inst{2}\orcidID{0000-0002-8086-8461} \and
Dickson Owuor\inst{3}\orcidID{0000-0002-0968-5742}} 
\authorrunning{D. Kerre et al.}
%
\institute{LIRMM, Univ Montpellier, CNRS, Montpellier, France\\
\email{anne.laurent@umontpellier.fr}\\
\and
IES, Univ Montpellier, CNRS, Montpellier, France\\
\email{Kenneth.Maussang@umontpellier.fr} \and 
SCES, Strathmore University, Nairobi, Kenya\\
\email{\{dkerre, dowuor\} @strathmore.edu}}
\maketitle              
\begin{abstract}
A well structured collection of the various Quantum Cascade Laser (QCL) design and working properties data provides a platform to analyze and understand the relationships between these properties. By analyzing these relationships, we can gain insights into how different design features impact laser performance properties such as the working temperature. Most of these QCL properties are captured in scientific text. There is therefore need for efficient methodologies that can be utilized to extract QCL properties from text and generate a semantically enriched and interlinked platform where the properties can be analyzed to uncover hidden relations. There is also the need to maintain provenance and reference information on which these properties are based. Semantic Web technologies such as Ontologies and Knowledge Graphs have proven capability in providing interlinked data platforms for knowledge representation in various domains. In this paper, we propose an approach for generating a QCL properties Knowledge Graph (KG) from text for semantic enrichment of the properties. The approach is based on the QCL ontology and a Retrieval Augmented  Generation (RAG) enabled information extraction pipeline based on GPT 4-Turbo language model. The properties of interest include: working temperature, laser design type, lasing frequency, laser optical power and the heterostructure. The experimental results demonstrate the feasibility and effectiveness of this approach for efficiently extracting QCL properties from unstructured text and generating a QCL properties Knowledge Graph, which has potential applications in semantic enrichment and analysis of QCL data.
\keywords{Information Extraction \and Knowledge Graphs \and Linked Data \and Ontologies \and Retrieval Augmented Generation \and Semiconductor Lasers \and Semantic Web}
\end{abstract}
\section{Introduction}
Quantum Cascade Lasers (QCL) are semiconductor laser devices which consists of nanometric stack of different semiconductor materials and whose spectral emission is restricted to the frequency range from about 100GHz to 10THz. The stacks of different materials are referred to as heterostructures. Interactions between the various layers of materials results to various emission behaviour of the QCL laser device\cite{ref_article1}. The nature of the radiations emitted by these laser devices have enabled various applications ranging from analysis of chemicals, high-resolution spectroscopy in astronomy, detection of organic compounds in drugs etc \cite{ref_article2,ref_article3,ref_article4,ref_article5,ref_article6,ref_article7}. 
\par QCL properties can be broadly classified into two categories: Design properties and the Optoelectronic/ Working properties. The design features consists of the laser design characteristics such as the laser design types, the material combinations used and the layer sequencing. The Optoelectronic properties on the other hand refers to the performance behaviour of a QCL device with particular design characteristics. Examples of working properties include Power, Temperature, Frequency etc. \par The working properties of a QCL device are dependent on the design features. This implies that there is a relationship between the laser design and working properties. Understanding the relationships between the QCL design and working properties plays an important role in the fabrication of a QCL semiconductor device with target properties, for instance the Working Temperature. 
\par The QCL properties are described in text where several device designs and there corresponding working properties are proposed. Some of the laser properties mentioned in the text are based on references in other text. Some of the QCL properties such as temperature may have several values in text hence its important to extract the correct value of interest. A sample text description of QCL properties is given: \textit{``GaAs/AlGaAs quantum cascade lasers based on four quantum well structures operating at 4.7 THz are reported. A large current density dynamic range is observed, leading to a maximum operation temperature of 150 K for the double metal waveguide device and a high peak output power more than 200 mW for the single surface plasmon waveguide device"}\cite{ref_article8}. This text description contains several QCL properties such as lasing frequency (4.7 THz), power (200 mW), working temperature (200 K) and the heterostructure materials (GaAs/AlGaAs). The data on QCL properties therefore exists in heterogenous text sources in an unstructured format and therefore requires a lot of effort to collect data, structure and explore the relationships between the various QCL properties. 
\par There has been attempts to extract QCL properties from text using rule-based approaches\cite{ref_article9}, indicating a possibility to generate QCL data from text. A formal representation of the QCL properties in form of an ontology model has also been proposed\cite{ref_article10}. The existing Knowledge bases in the materials science domain don't capture properties for the QCL domain and cannot therefore be readily utilized to answer queries on QCL design features and the corresponding performance characteristics. 
\par There is therefore the need for a semantically enriched platform that captures the QCL properties in text,  their provenance information together with links to the references for the properties. This will enable exploration of the relationships between the various properties in form of queries. The insights derived from these information can be used in the fabrication of laser devices with target properties. This  will also provide an interlinked platform where both
machines and humans can explore and query the QCL properties data in a FAIR (Findable, Accessible, Interoperable, and Reusable) manner\cite{ref_article11}.  

\par In this paper, we present an original methodology for semantic enrichment of QCL properties in text via a Knowledge graph generation from text. The main contributions of this paper are therefore as follows:
\begin{enumerate}[label=(\roman*)] 
\item We propose a Retrieval Augmented Generation (RAG) based approach for QCL property extraction from text.
\item We present an experimental analysis of our approach on various Large Language Models on QCL property Extraction from Text.
\item We generate a Knowledge Graph of QCL properties from scientific articles for the semantic enrichment of QCL properties.
\item We evaluate the ability of the generated Knowledge Graph in capturing knowledge for properties in the QCL domain. 
\end{enumerate}
\par The remaining sections of this paper as organized as follows: we give an overview of related works in information extraction and knowledge representation in the materials science domain in section ~\ref{sec:rw}, the Knowledge Graph generation workflow in section ~\ref{sec:method}, the experimental evaluation in section ~\ref{sec:exp}, the results and discussions in section ~\ref{sec:results} and lastly conclude in section ~\ref{sec:conc}.
\section{Related Work \label{sec:rw}}
\subsection {Information Extraction in Materials Science Domain}
Extraction of materials properties from text has gained a lot of interest recently as occasioned by the need for accelerated materials discovery. The methodologies developed can be broadly classified in to rule-based methods and machine learning based methods. Rule-based methods constitutes rules, grammars and other expert-defined structures for identification of properties in specific domains. The machine learning based approaches entails training of learning algorithms on labelled data to enable them to learn how to identify specific materials properties in text.
\par Examples of rule based toolkits adopted for properties extraction in materials science include chemDataExtractor\cite{ref_article12}, LeadMine\cite{ref_article13}, ChemicalTagger\cite{ref_article14}, tmChem\cite{ref_article15} and ChemSpot\cite{ref_article16}. The chemDataExtractor toolkit has also been widely adopted for materials properties extraction in other specific use cases which includes: thermo-electric materials\cite{ref_article17}, semiconductor bandgaps\cite{ref_article18}, refractive indices and dielectric constants\cite{ref_article19}, an auto-populated ontology of materials science\cite{ref_article20}, battery materials\cite{ref_article21}, transition temperatures of magnetic materials\cite{ref_article22} and quantum cascade laser properties\cite{ref_article9}. 
\par Machine learning methods have also been adopted in the extraction of materials science properties. Examples include generation of datasets of gold nano-particle synthesis procedures, morphologies and size entities\cite{ref_article23}, and materials synthesis recipes\cite{ref_article24}.  Another work is on the use of the combination
of deep convolutional and recurrent neural networks for named entity recognition\cite{ref_article25}. 
BERT(Bidirectional Encoder Representations from
Transformers) models have also been proposed for the analysis of optical materials\cite{ref_article26} and extraction of battery
materials from scientific text\cite{ref_article27}. \par The emergence of generative large language models have also opened opportunities in the extraction of materials properties from text. The models have been harnessed for text parsing in solid-state synthesis internary chalcogenides\cite{ref_article28} .The in-context learning method is also applied to assess the ability of LLMs in processing materials data\cite{ref_article29} and extracting materials data from research papers\cite{ref_article30}. Lastly, LLMs have also been utilised in the construction of functional materials Knowledge Graph in multidisciplinary materials science\cite{ref_article31}. 
\subsection{Knowledge Graphs in Materials Science Domain}
Knowledge Graphs have been proposed in representing knowledge for properties in the materials science domain. The KGs are based on several foundational and specific domain ontologies developed for this domain. The motivation behind these KGs is to provide an accelerated analysis of materials science properties for various reasons, including materials discovery. The proposed KGs
ranges from specific materials such as nanocomposite materials\cite{ref_article32} and a wide range of materials in this domain\cite{ref_article33,ref_article34,ref_article35,ref_article36,ref_article37}. A Knowledge Graph for materials experiment has also been proposed\cite{ref_article38}. The adoption of KGs in the materials science domain underscore the impact of this technologies in the materials science domain. The development of the QCL ontology also points on an opportunity for the representation and analysis of the QCL properties\cite{ref_article10}.     
\subsection{Summary}
Despite the great advancements in Information Extraction in materials science domain, there are a couple of problems to be addressed: the rule-based approach developed for QCL property extraction from text is specific and is limited in cases where their is slight change in text structure. The machine learning models need quality training data to effectively train and evaluate the models for this task. The LLM based methodologies gives a promising direction in the extraction of QCL properties from text. The proposed KGs in the materials science domain do not capture the properties of interest in the QCL domain. They cannot therefore be adopted for QCL properties representation and exploration. \par 
There is therefore need for efficient methodologies for extracting QCL properties from text to generate structured data and utilise this data to generate a Knowledge Graph for representing QCL properties, relationships among them and the provenance information. To the best of our knowledge, this work presents the initial steps for implementing the task of QCL properties extraction from text and KG generation for property exploration.
\section{Methodology \label{sec:method}}
The Knowledge Graph generation approach is composed of the following parts: Information Extraction pipeline and the KG modelling and data enrichment part. 
We describe these parts in the following subsections:
\subsection{Information Extraction Pipeline \label{sec: IE}}
\begin{figure}
\centering
\includegraphics[width=1.0\textwidth]{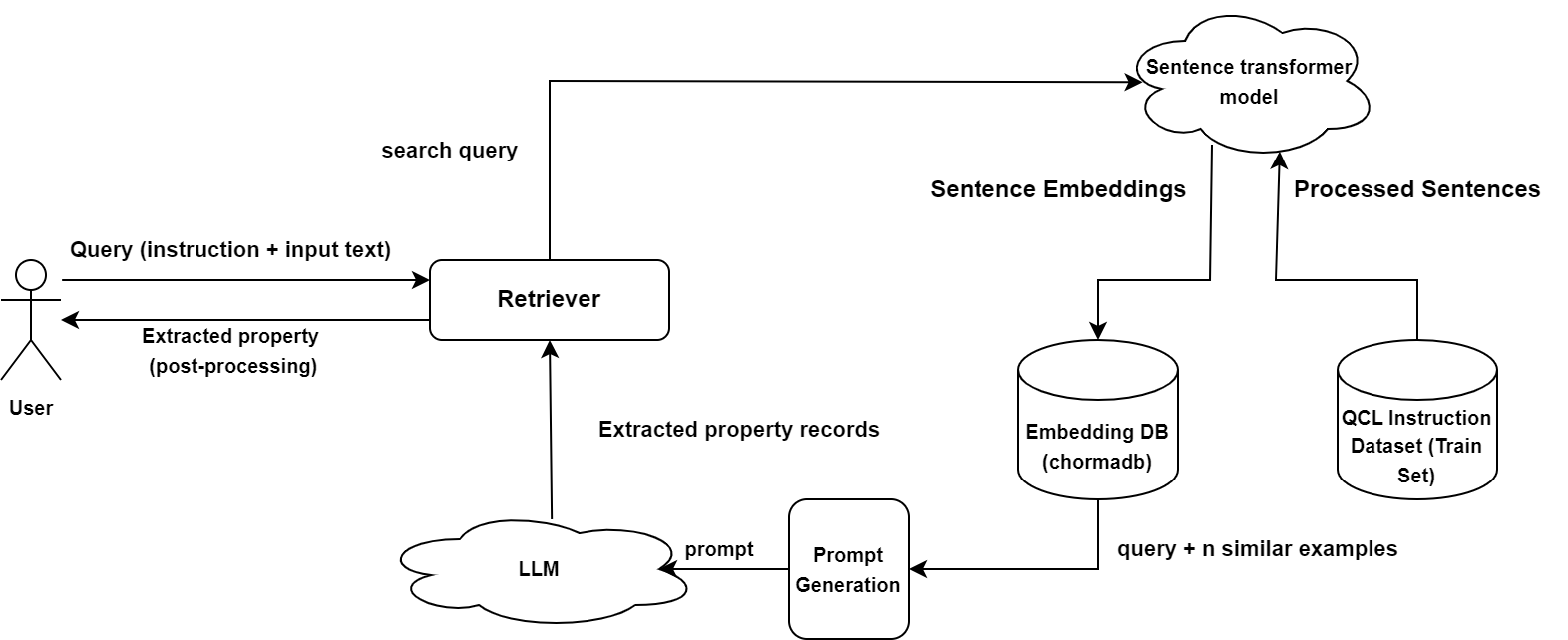}
\caption{The Advanced RAG Pipeline} \label{fig:RAG_pipeline}
\end{figure}

In this module, we explore the use of large language models in developing the pipeline. We approach this process as an open information extraction task as we are interested in literal values for the properties i.e a value and a unit. We propose a RAG enabled pipeline based on GPT-4 Turbo. This is owed to GPT-4 Turbo improved efficiency in generating responses and the larger context window ~\cite{ref_article39}.  Our approach is based on advanced RAG~\cite{ref_article40}. Large language models such as GPT models are trained on general knowledge data and cannot be efficiently used on specific domain tasks without adaptation. We hypothesize that exposing the model to labelled data consisting of sample text describing QCL properties and the corresponding extracted properties improves the model's performance on this task of property extraction from text. This also aligns the model's output to the expected format and minimizes irrelevant responses.  As illustrated in Figure ~\ref{fig:RAG_pipeline}, the module has three sections i.e Retrieval, Data Augmentation and Data Generation. The rest of this subsection describes the pipeline modules.
\subsubsection{Retrieval:} An input sentence (query) containing QCL property of interest and an instruction to the large language model is submitted by the user to the retriever. This is then forwarded by the retriever to the data augmentation module for retrieval of similar responses based on the query.
\subsubsection{Data Augmentation:} This module Consists of train data embeddings stored in an embedding database (DB). This provides the context to the model during information extraction. In our case, the context is provided by an original QCL properties instruction dataset  \cite{ref_article41}. The dataset description is given elsewhere\cite{ref_article42}. We sample 80 \% of this dataset for training and 10 \% for testing. The dataset comprises of 1040 sample sentences containing QCL properties, an instruction to the model for information extraction together with the corresponding properties extracted.  Sentence embeddings are computed using a based pre-trained sentence transformer model\cite{ref_article43}. We adopt the  all-mpnet-base-v2\footnote[1] {\url{https://huggingface.co/sentence-transformers/all-mpnet-base-v2}} version of the sentence transformer model in our approach. The query embeddings are also computed by the sentence transformer model in order to allow for comparison with the emdeddings in the training data. \par
\begin{figure}
\centering
\includegraphics[width=0.85\textwidth]{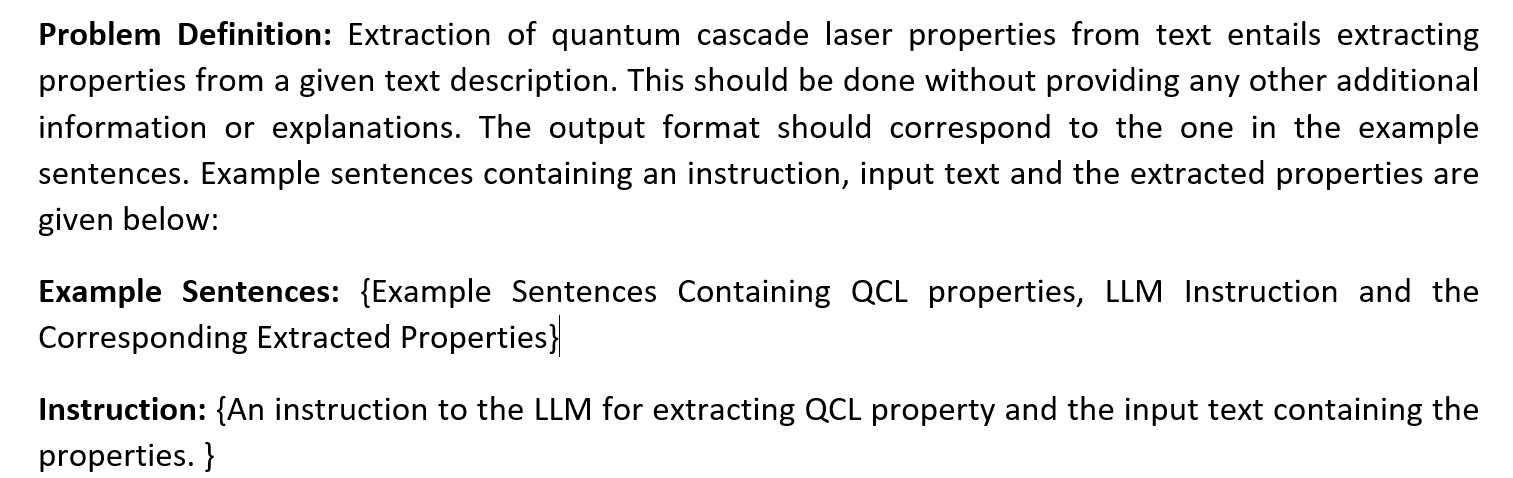}
\caption{The Prompt Template} \label{fig:prompt_template}
\end{figure}

 Similarity scores between the query embeddings and the train sentence emdeddings in the embedding DB are computed using the cosine similarity metric. The examples in the embedding DB that are more similar are retrieved. The examples and the user query are both  passed to a prompt generator which prepares a prompt based on a defined prompt template. We define a prompt template that allows parsing a user query together with the relevant examples showing sample instructions with text containing properties and the extracted properties. The generated prompt is then passed to the Generation phase. Figure ~\ref{fig:prompt_template} shows the prompt template and Figure ~\ref{fig:renegerated_prompt} shows a sample regenerated prompt.
\subsubsection{Generation:}
In the generation module, a response is generated by the language model in a zero shot manner. This stage is flexible as any language model can be used for response generation. The model responses are then processed to remove any incomplete records or any irrelevant responses. 

Formally, the general process of extraction of QCL properties from text based on the detailed RAG approach is carried out as follows: we frame the QCL property extraction from text task in the form of a conversation as
follows: Given a set of input sentences S, where S= \{s1, s2, s3...Sn\}, we design a prompt P that contains an instruction I, the Sentences S and contextual examples C= \{C1,C2,C3..Ck\},  where k depends on the number of examples desired for the context. This prompt is passed to the model in order to extract a particular QCL property record R, where R= \{p1,p2,p3...pn\} and n implies the number of properties in the record. 
Algorithm \ref{alg:one} shows the steps followed to extract the data.
\RestyleAlgo{ruled}
\SetKwComment{Comment}{/* }{ */}
\begin{algorithm}
\caption{The QCL Properties Extraction Steps}\label{alg:one}
\KwIn{Set of input sentences S, Prompt P, Instruction I, base model M, an embedded instruction dataset to provide the context C.}
\KwOut{QCL property record R.} 

Set the number of examples to be retrieved (k) to a specific value. 
 \Comment{This will determine the number of examples to be retrieved from the context documents based on the user query.} 

Enter an instruction I and an input sentence S.

Set the query Q= Instruction (I) + Input Sentences (S).

Convert the queries to a vector embedding Eq.

Pass query Eq to the retriever.

Fetch k examples from C based on Eq.

Set Eq = Eq + k examples.

Set Q= Eq (decoded to plain text). 

Update the prompt P with Q.

Pass P to the base model M. 

\Return {R} 

\textbf{Repeat} steps 2-9 until all the property records of interest are extracted. 

Post-process R. \Comment{Entails removal of incomplete records and any unnecessary responses.}

\end{algorithm}

The pipeline is used to extract QCL properties data from 42 open access abstracts containing the various QCL properties for 42 laser devices \cite{ref_article8,ref_article44,ref_article45,ref_article46,ref_article47,ref_article48,ref_article49,ref_article50,ref_article51,ref_article52,ref_article53,ref_article54,ref_article55,ref_article56,ref_article57,ref_article58,ref_article59,ref_article60,ref_article61,ref_article62,ref_article63,ref_article64,ref_article65,ref_article66,ref_article67,ref_article68,ref_article69,ref_article70,ref_article71,ref_article72,ref_article73,ref_article74,ref_article75,ref_article76,ref_article77,ref_article78,ref_article79,ref_article80,ref_article81,ref_article82,ref_article83,ref_article84}. An abstract documenting a QCL property captures all the properties mentioned for a particular QCL device. The data is post-processed to have a clean file of all the properties extracted. We also include the metadata (DOI and URL) for provenance information and references for referencing the various mentioned properties during the post-processing phase. The missing values are also included in the post-processing stage. The final data constitutes a well structured csv file containing QCL properties data for every device  together with the associated provenance information and the references.
\subsection{Knowledge Graph Modelling and Data Enrichment}
In this section, there are two processes that we carry out: first we define the KG model to organize the data and secondly we map the data to enrich it. We detail them in the following subsections:
 \newline \newline
\textbf{The Knowledge Graph Modelling:}
A Knowledge Graph represents a semantic network of interlinked entities. Entities refers to  real world concepts or ideas that can be identified by a unique identifier on the web. A Knowledge Graph is defined in form of triples, that consists of two entities and a relation (predicate) linking them. Formally, a Knowledge graph KG can be defined as KG = \{t1, t2, t3...tn\} and t refers to the various triples in the KG and n the number of triples in the KG. A triple t= \{s,p,o\} where s is the subject, p the predicate and o the object. s, p and o are denoted by Resource Identifiers inform of Uniform Resource Identifiers (URIs) or Literals (e.g., strings, numbers, dates). The semantics of a KG are provided by ontologies or standard vocabularies. \par
\begin{figure}[htb]
\centering
\includegraphics[width=1.07\textwidth]{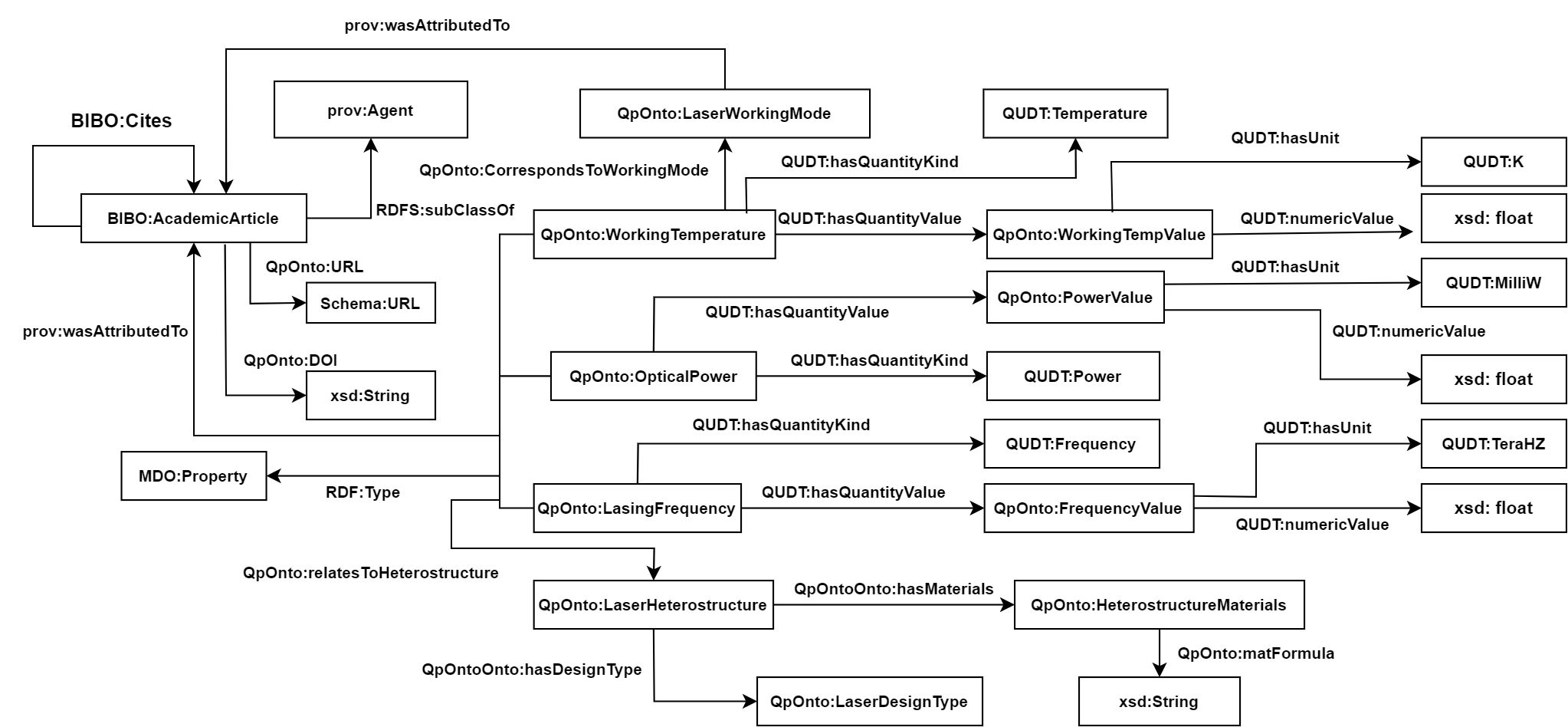}
\caption{The Knowledge Graph Schema} \label{fig:KG-Schema}
\end{figure}

In our case, the entities are the various QCL properties, the various relationships among them and the provenance information for these properties. We define the KG schema whose semantics are provided by the QCL ontology model\footnote[2]{\url{https://github.com/DeperiasKerre/qcl_Onto/blob/main/qclontology/version-1.0/qclonto.owl}}  \cite{ref_article10} and other vocabularies such as BIBO\footnote[3] {\url{https://dcmi.github.io/bibo/}} and schema.org\footnote[4]{\url{https://schema.org/}}. Figure ~\ref{fig:KG-Schema} shows the KG schema used to organize the data and table ~\ref{tab1} shows the prefixes and URIs for the namespaces used in the KG schema..
\newline \newline
\begin{table}
\centering
\caption{Namespace URIs}\label{tab1}
\begin{tabular}{|p{2.8cm}|p{11.0 cm}|}
\hline
{\bfseries Prefix} & {\bfseries Namespace URI} \\
\hline
QpOnto & \url{https://github.com/DeperiasKerre/qcl_Onto/blob/main/qclontology/version-1.0/qclonto.owl#}\\ 
MDO & \url{https://w3id.org/mdo/core/}
\\
BIBO & \url{https://dcmi.github.io/bibo/#:}\\
prov & \url{http://www.w3.org/ns/prov#}\\
RDFS & \url{http://www.w3.org/2000/01/rdf-schema#}\\
QUDT\_Properties & \url{https://qudt.org/schema/qudt/}\\
QUDT\_Units & \url{https://qudt.org/vocab/unit/}\\
QUDT\_Quantity Kinds & \url{https://qudt.org/vocab/quantitykind/}
\\
\hline
\end{tabular}
\end{table}
The KG schema covers the following categories of information that we enrich: laser heterostructure (heterostructure materials), laser working properties (power, temperature and the lasing frequency), laser design types, provenance information and citation tracks. The heterostructure captures the materials stacking properties of the laser and the related design. It is defined by the concept QpOnto:LaserHeterostructure. A heterostructure contains heterostructure materials (QpOnto:HeterostructureMaterials) and the material combination has a formula (QpOnto:matFormula) which indicates the materials used and the ratio of combination in a string. A heterostructure also has a design type. Examples of design types includes the resonant phonon and the LO phonon design types. \par The optoelectronic properties captures the QCL performance behaviour as a result of injection of current. These includes the working temperature, power and frequency. This properties are related to particular design information i.e design types and heterostructure materials. The working temperature also depends on the laser working mode i.e whether the emission is in continuous or pulse mode. \par
\begin{figure}
\scriptsize
\centering
\includegraphics[width=1\textwidth]{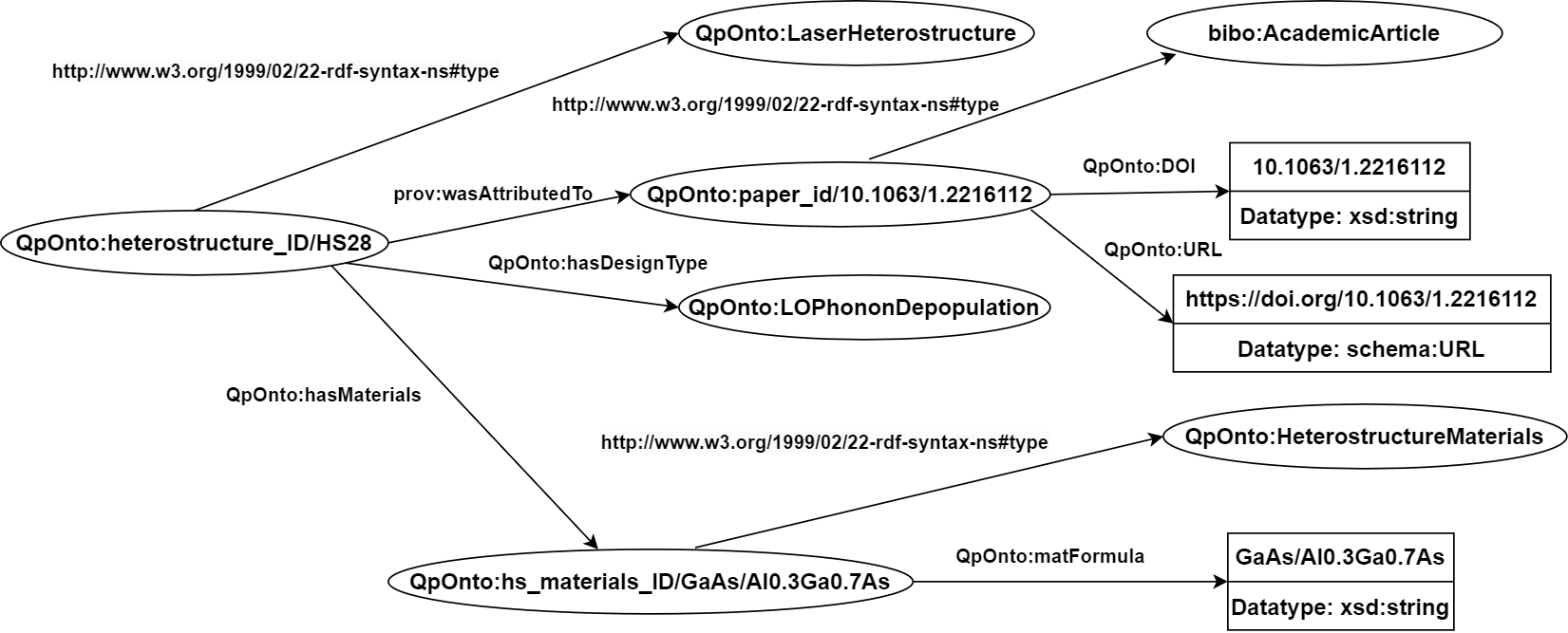}
\caption{Sample Instance of the QCL Heterostructure Visualization from the Generated Knowledge Graph} \label{fig:hs_instance}
\end{figure}

 The QCL properties provenance information is also captured in the schema. This is implemented by capturing the metadata (DOI and URL) of the articles documenting the various laser properties. We also provide links to references in those articles as some of the properties proposed in a given design are based on another design in a referenced article. This is done using the concept BIBO:cite from the BIBO ontology. The semantics of the units, quantity kinds, numerical values and their associated relationships for the QCL working properties are modelled by re-using the terms in the QUDT ontology. 
\newline \newline \textbf{Data Mapping:} In this phase, several steps are carried out in order to perform mapping of the data values generated in section \ref{sec: IE} to the entities. This is implemented via the data properties. All the object properties are also implemented. The the KG classes are also instantiated.  We also specify the data types for all the data instances. This is implemented using the rdflib library \footnote[5]{\url{https://github.com/RDFLib/rdflib}}. The units, design types and the relevant working modes are also enriched with the relevant URIs. The redundant triples are also examined and eliminated from the generated Knowledge Graph. The final RDF file is then serialized into the Turtle and the RDF/XML formats in order to generate the Knowledge Graph for querying and exploration. The generated Knowledge Graph contains a total of 3403 triples containing the QCL properties and their associated provenance information. A visualization of a sample  instance of a QCL heterostructure (capturing the design type and material combination information) and its provenance information is shown in Figure \ref{fig:hs_instance}.
%
\section{Experiments \label{sec:exp}}
In this section, we carry out experiments to evaluate the KG generation approach. This is done is based on three strategies: the performance of the approach in QCL property extraction from text, the consistency and the correctness of the generated KG in terms conformance to the QCL properties domain requirements i.e the ability to capture the intended knowledge correctly. 
\subsection{Property Extraction From Text}
\subsubsection{Evaluation Metrics:} For the evaluation of the information extraction module, we adopt the expert validation approach. This entails comparison of the model's output with the expert annotated ground truth label in the evaluation dataset. We use the precision and recall in order to evaluate the performance  of the approach on QCL property extraction from text\cite{ref_article85}. 
Precision is the fraction of correct (relevant) records among all extracted records and the recall is the fraction of successfully extracted records among all correct (relevant) records in the dataset. The metrics are determined as follows:
\begin{equation}
    Precision = \frac{TP}{TP + FP}
\end{equation}

\begin{equation}
    Recall = \frac{TP}{TP + FN}
\end{equation}

TP refers to the true positive count (the number of  correct records extracted), FP is the false positive count (the number  incorrect records extracted), and FN corresponds to the false negative count (the number of correct records that are not extracted).
We define the terms correct and incorrect in our context as follows: \newline \newline \textbf{Definition 1:} The word “correct” in this context implies that the property value extracted can be validated by a human expert when reading the corresponding sentence containing the property. It should also match with the ground truth label value in the evaluation dataset. Property values with units are only considered correct if they are extracted together with the units. \newline \newline \textbf{Definition 2:} An “incorrect” (false)
record suggests that the value extracted does not correspond to the correct/expected value as compared to the ground truth value in the evaluation dataset.

\par
  \subsubsection{Evaluation Dataset and Baselines:} We evaluate the information extraction pipeline on a test dataset obtained from the instruction dataset described in section ~\ref{sec: IE}. This contains a total of 130 sentences containing the different QCL properties. Table ~\ref{tab2} gives the summary statistics of the test data per QCL property.
  
\begin{table}[htb]
\centering
\caption{Statistics of the test data.}\label{tab2}
\begin{tabular}{|p{5.9cm}|p{3.5 cm}|}
\hline
{\bfseries Quantum Cascade Laser Property} & {\bfseries Number of Sentences}\\
\hline
Laser Optical Power & 26 \\
Working Temperature & 34 \\
Laser Design Type  & 27 \\
Laser Frequency & 29\\
Laser Heterostructure & 14 \\
\textbf{Total} & \textbf{130}\\
\hline
\end{tabular}
\end{table}
For the baseline, we evaluate our approach on GPT 4-Turbo and the Mistral-7b-Instruct model. We don't perform evaluations on GPT-4 due to the context window limitations. We compare the performance of these models on QCL property extraction from text with and without (simple query) the proposed RAG approach. We analyze the performance per QCL property as different properties have varying levels of difficulty during the extraction process. 
\subsection{Knowledge Graph Evaluation}
The generated Knowledge Graph is evaluated based on two metrics: the consistency in the KG triples and the correctness of the KG in capturing the intended knowledge in the QCL domain properties. The consistency ensures the logical soundness of the defined triples in the generated KG. The correctness of the KG in terms of the domain requirements aims to evaluate the generated KG`s ability in capturing the domain knowledge of interest and being able to provide answers to questions regarding the avrious QCL properties. \par In order to validate the ability of the KG to conform to the domain requirements, we define a set of 7 classes of competency questions (CQs) with the help of the QCL domain experts. The competency questions capture the whole QCL properties of interest i.e the design, working properties,  the laser working modes and the provenance information. The competency questions entails probable questions that the QCL experts could be interested to get answers to when trying to understand the implications of the laser design on the overall laser performance.  Some of the competency questions are adopted from the user requirements of the QCL ontology. Table ~\ref{tab3} shows the classes of the CQs.
\begin{table}
\centering
\caption{List of Competence Questions}\label{tab3}
\begin{tabular}{|p{1.5cm}|p{11.2 cm}|}
\hline
{\bfseries Question ID} & {\bfseries Question Text}\\
\hline
CQ1 & What are the possible material compositions of a QCL laser heterostructure with a design type X ?\\
CQ2 & What is the working property X of a QCL laser working in mode Y ?\\
CQ3 & What is the performance property X of a QCL laser having a heterostructure with material composition Y?\\
CQ4 & For a particular performance property X, what are the corresponding laser heterostructure designs?\\
CQ5 & For a particular performance property X, what are the corresponding heterostructure material compositions?\\
CQ6 & What are  the the DOIs and/ the URLs of the scientific articles documenting a laser with performance property W or with heterostructure materials X or working mode Y or design type Z. \\
CQ7 & What are the DOIs and URLs of the articles being referenced by a QCL device with property W or with heterostructure materials X or working mode Y or design type Z?\\
\hline
\end{tabular}
\end{table}
The X,Y and Z are place holders for any particular properties of interest to be used in the  queries. For every class of queries in Table \ref{tab3}, we design and run several possibilities of the queries. We compare the queries output with the expected output in order to determine the precision in question answering by the Knowledge Graph. 
\section{Results and Discussions \label{sec:results}}
In this section, we present and discuss the results for the evaluation of the KG generation approach based on the property extraction from text and the ability of the generated KG to meet the QCL domain requirements. We detail them in the following subsections:
\subsection{Property Extraction from Text}
We present the results for QCL property extraction from text analyzed per QCL property as the properties have a varying level of difficulty during the extraction process. The results are as shown in Table \ref{tab4} for the laser power, Table \ref{tab5} for working temperature, Table \ref{tab6} for laser heterostructure materials, Table \ref{tab7} for lasing frequency, Table \ref{tab8} for the laser design type and Table \ref{tab9} for the average performance of the models for all the properties. 
\begin{table}[htb]
\scriptsize
\centering
\caption{Performance Metrics for Laser Output Power.}\label{tab4}
\begin{tabular}{|p{6.0cm}|p{2.5cm}|p{2.5 cm}|}
\hline
{\bfseries Model} & {\bfseries Precision} & {\bfseries Recall} \\
\hline
gpt 4-turbo + RAG & 100 \% & 100 \% \\
gpt 4-turbo (simple query) & 100 \% & 100 \% \\
mistral-7b-instructv0.3 + RAG & 100 \% & 100 \%\\
mistral-7b-instructv0.3 (simple query) & 100 \% & 96.15 \%\\
\hline
\end{tabular}
\end{table}
 
\begin{table}[htb]
\scriptsize
\centering
\caption{Performance Metrics for the Laser Working Temperature.}\label{tab5}
\begin{tabular}{|p{6.0cm}|p{2.5cm}|p{2.5 cm}|}
\hline
{\bfseries Model} & {\bfseries Precision} & {\bfseries Recall} \\
\hline
gpt 4-turbo + RAG & 100 \% & 100\% \\
gpt 4-turbo (simple query) & 90.91 \% & 96.77 \% \\
mistral-7b-instructv0.3 + RAG & 85.30\% & 100 \%\\
mistral-7b-instructv0.3 (simple query) & 85.29 \% & 100 \%\\
\hline
\end{tabular}
\end{table}

\begin{table}[htb]
\scriptsize
\centering
\caption{Performance Metrics for the Laser Heterostructure.}\label{tab6}
\begin{tabular}{|p{6.0cm}|p{2.5cm}|p{2.5 cm}|}
\hline
{\bfseries Model} & {\bfseries Precision} & {\bfseries Recall} \\
\hline
gpt 4-turbo + RAG & 85.71 \% & 100 \% \\
gpt 4-turbo (simple query) & 91.67 \% & 84.62 \% \\
mistral-7b-instructv0.3 + RAG & 85.71\% & 85.71 \%\\
mistral-7b-instructv0.3 (simple query) & 92.86 \% & 100 \%\\
\hline
\end{tabular}
\end{table}
\begin{table}[htb]
\scriptsize
\centering
\caption{Performance Metrics for the Laser Frequency.}\label{tab7}
\begin{tabular}{|p{6.0cm}|p{2.5cm}|p{2.5 cm}|}
\hline
{\bfseries Model} & {\bfseries Precision} & {\bfseries Recall} \\
\hline
gpt 4-turbo + RAG & 100\% & 100\% \\
gpt 4-turbo (simple query) & 92.86 \% & 96.30 \% \\
mistral-7b-instructv0.3 + RAG & 96.43 \% & 96.43 \%\\
mistral-7b-instructv0.3 (simple query) & 92.59 \% & 92.59 \%\\
\hline
\end{tabular}
\end{table}
\begin{table}[htb]
\scriptsize
\centering
\caption{Performance Metrics for the Laser Design Type.}\label{tab8}
\begin{tabular}{|p{6.0cm}|p{2.5cm}|p{2.5 cm}|}
\hline
{\bfseries Model} & {\bfseries Precision} & {\bfseries Recall} \\
\hline
gpt 4-turbo + RAG & 100 \% & 33.33 \% \\
gpt 4-turbo (simple query) & 92.31 \% & 46.15 \% \\
mistral-7b-instructv0.3 + RAG & 100\% & 92.60 \%\\
mistral-7b-instructv0.3 (simple query) & 33.33 \% & 29.41 \%\\
\hline
\end{tabular}
\end{table}
\begin{table}[htb]
\scriptsize
\centering
\caption{Average Performance of the Models for all the Properties}\label{tab9}
\begin{tabular}{|p{6.0cm}|p{2.5cm}|p{2.5 cm}|}
\hline
{\bfseries Model} & {\bfseries Precision} & {\bfseries Recall} \\
\hline
gpt 4-turbo + RAG & \textbf{97.14 \%} & \textbf{86.67 \%} \\
gpt 4-turbo (simple query) & 93.55 \% & 83.59 \% \\
mistral-7b-instructv0.3 + RAG & 93.49 \% & 92.01 \%\\
mistral-7b-instructv0.3 (simple query) & 80.82 \% & 80.69 \%\\
\hline
\end{tabular}
\end{table}

The laser power, working temperature and frequency exhibit higher precision in the extraction process in both simple query and the proposed approach. This is attributed to the fact that these properties are generally available in the general knowledge of the models. However, their is lower recall for the simple queries for these properties. This is in cases where more than one value of these properties are mentioned and the models struggle to identify the required value hence failing to extract these properties. This is also the case when the values are given in ranges, for instance, the lasing frequency. Provision of examples capturing such scenarios improves the model performance on these properties.  This is indicated by the  best performance exhibited by the GPT 4-Turbo    RAG based approach as shown in Tables \ref{tab4}, \ref{tab5} and \ref{tab7}. It is also noted that there is a significant improvement in performance for both the precision and recall for the lasing frequency and the working temperature with the RAG based approach(Tables \ref{tab5} and \ref{tab7}). This is due to the ability of the model to learn how to identify the properties of interest based on the provided examples. \par The laser design type property entails a QCL domain specific property. The models exhibits higher precision in cases where the keyword ``design type"  is explicitly used in the text description but fails totally to recognize and extract the property in cases where the key word is not used. For both models, there is an improvement in generalization (in terms of precision) as the models utilize the labelled examples to recognize this domain specific terms. There is however a decrease in recall for the GPT-Turbo model. This is attributed to cases where the model does not completely extract terms with more variations form the context documents. In this case, the Mistral-instruct-7b model exhibits the best performance with the proposed RAG approach as shown in Table \ref{tab8}. \par The QCL heterostructure materials property is also a QCL domain specific property. With this property, there is no significant improvement with the proposed approach for all the models. Despite the heterostructure being a domain specific concept, its description is characterized by the terms```structure", ``heterostructure" or ``materials" which enables the models to identify the properties at relatively higher precision even with simple query. The best performance is exhibited by the mistral 7b-instruct model in a simple query format (Table \ref{tab6}). 
\par In summary, exposure of large language models to quality labelled data improves their ability in recognizing and extracting the relevant QCL properties. This is not however the case with all the properties, for instance the laser heterostructure. With the proposed approach, its even possible to train the model to adopt a certain output format for the extracted properties to avoid any unnecessary
responses or undesired output formats. This approach can be extended for the other QCL properties as it enables the model to learn how to identify the domain specific properties with the provided examples with less resources. With this approach, the models performance is however dependent on data quality and the properties covered. The models performance therefore increases with more diverse examples in the model's context.             
\subsection{Knowledge Graph Evaluation}
The Knowledge Graph consistency is validated by lack of inconsistencies/contradictions in the generated triples. We the validate the KG consistency using the pellet reasoner \cite{ref_article86}. For the Knowledge Graph correctness, We run a total of 20 queries in order to validate the suitability of the generated KG in capturing the QCL properties, the relationships between them and their provenance information. Table \ref{tab10} shows the specific queries run for each class of queries specified in Table \ref{tab3}. 
\par
\begin{figure}
\centering
\includegraphics[width=1\textwidth]{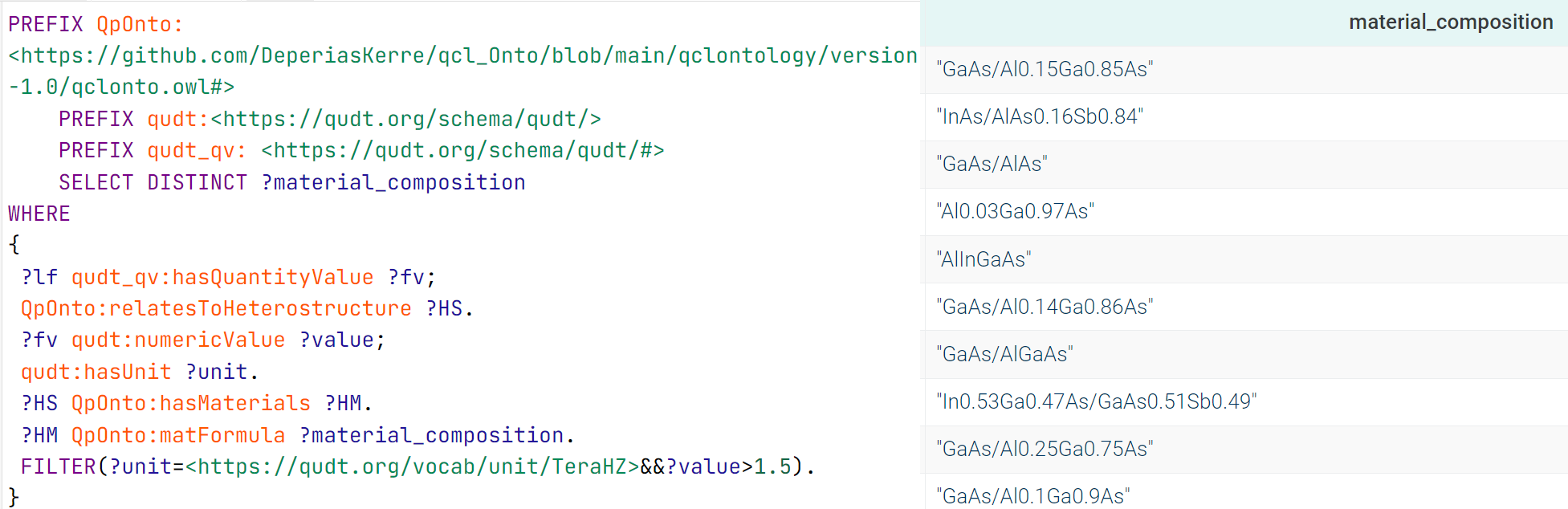}
\caption{A SPARQL Query and the Results for Query 5.2}\label{fig:query}
\end{figure}

 The queries range from simple to complex queries regarding the QCL properties and their provenance information.   
\begin{table}[htb]
\centering
\scriptsize
\caption{The Specific Queries Implemented for each Query Class}\label{tab10}
\begin{tabular}{|p{2.4cm}|p{11.0 cm}|}
\hline
{\bfseries Query Number} & {\bfseries Query Text}\\
\hline
1.1 & What are the possible material compositions of a QCL laser heterostructure with an LO Phonon Design Type ?\\
1.2 &  What are the possible material compositions of a QCL laser heterostructure with a Resonant Phonon Design Type?\\
1.3 & What are the possible material compositions of a QCL laser heterostructure with a Bound to Continuum design type?
\\
2.1 & What are the working temperatures for a QCL laser operating in the continuous wave mode?\\
2.2 & What are the working temperatures for a QCL laser operating in the pulsed mode?
\\
3.1 & What are the possible power values for a QCL laser with a heterostructure with material composition GaAs/Al0.15Ga0.85As?
\\
3.2 & What are the possible frequency values for a QCL laser with a heterostructure with material composition In0.53Ga0.47As/GaAs0.51Sb0.49?
\\
3.3 & Query 3.3: What are the possible working temperature values for a QCL laser with a heterostructure with material composition GaAs/Al0.3Ga0.7As?\\
4.1 & Query 4.1: What are the possible heterostructure designs for a QCL device with a working temperature greater than 100 K in Pulsed Mode?\\
4.2 & What are the possible heterostructure designs for a QCL device with an optical power less than 50 mW?
\\
5.1 & What are the possible heterostructure material compositions for a QCL device with a working temperature less than 85 K in the contionous wave mode? 
\\
5.2 & What are the possible heterostructure material compositions for a QCL device with a lasing frequency greater than 1.5 THz?
\\
6.1 & What are the DOIs and URLS of scientific articles documenting QCL laser devices with an optical power greater than 10mW?
\\
6.2 &  What are the DOIs and URLS of scientific articles documenting QCL laser devices with a working temperature greater than 100 K in pulse mode?
\\
6.3 & What are the DOIs and URLs of scientific articles documenting a QCL laser with a material composition of GaAs/Al0.25Ga0.75As?
\\
6.4 & What are the DOIs and URLs of scientific articles documenting QCL lasers with bound to continuum design type?
\\
6.5 & What are the DOIs and URLs of articles documenting QCL lasers with a heterostructure of material composition GaAs/Al0.15Ga0.85As, LO phonon design type and working temperatures greater than 70 K in pulse mode operation?
\\
7.1 & What are the DOIs and URLs of the articles being referenced by a QCL device with a working temperature greater than 225 K in the continuous wave mode?
\\
7.2 & What are the DOIs and URLs of the articles being referenced by a QCL device with an optical power less than 1 mW?
\\
7.3 & What are the DOIs and URLs of the articles being referenced by a QCL device with a lasing frequency greater than 2.5 THz and an LO Phonon design type?\\
\hline
\end{tabular}
\end{table}
We present a scenario where a QCL expert is interested in the heterostructure materials composition of a QCL device with a certain working property, for instance a frequency value greater than 1.5. Such a question can be captured by query 5.2 in Table \ref{tab10}. A corresponding SPARQL query and the retrieved results for query 5.2 are shown in  Figure \ref{fig:query}. 
All the queries are successfully answered by the generated KG and the complete results are available in the GiHhub repository \footnote[6]{\url{https://github.com/DeperiasKerre/qKG}}. \par The ability of the generated KG in successfully answering the competency questions indicates its capability in capturing QCL properties information from the various textual sources. This provides a unified platform that allows exploration of QCL properties in a structured manner to be able to derive insights on the relationship between the properties as opposed to manually exploring the textual documents for these properties. This is useful in scenarios where there is need for a quicker comparison of the various QCL properties for instance, the working properties for  a particular laser design. The ability to capture the provenance information for the QCL properties also makes it possible to track the sources of this information via permanent identifiers such as the DOI and the URL. With the generated KG, it is also possible to have a linked acess to the references for the various QCL properties as some of the properties are based on other properties mentioned in the references. This therefore provides an efficient way of accessing this information for quicker analysis. 
\section{Conclusions and Future Work \label{sec:conc}}
In this paper, we address the issue of semantic enrichment of QCL properties in text by presenting an approach for generating a Knowledge Graph for QCL properties from text. This enables exploration of the properties in the various heterogenous textual sources while maintaining provenance information. The approach is composed of two parts: an information extraction pipeline for extracting QCL properties from text based on an LLM enabled RAG approach and the data enrichment part where all the data is mapped and the relationships interlinked. The semantics of the QCL properties KG are provided by the QCL ontology and other vocabularies. We evaluate the KG generation approach based on two strategies i.e the performance in QCL property extraction from text and the correctness of the generated Knowledge Graph in modelling the knowledge in the QCL properties domain. \par The proposed information extraction approach presents competitive results indicating the model's ability to learn how to identify domain specific properties with the help of curated examples.  The generated Knowledge Graph indicates its ability in modelling the knowledge in the QCL properties and their provenance information hence providing a semantically enriched, unified platform for quicker analysis and insights regarding the fabrication of QCL laser devices with target properties. Overally, our work represents an important step towards the development of automated methods for extracting and representing complex scientific knowledge regarding QCL properties from text. We believe that this approach has the potential to transform the way that researchers interact with scientific literature capturing QCL properties, and open up new avenues for discovery and innovation in QCL device fabrication.
\par The generated Knowledge Graph  is however based on a limited number of articles and properties. This can be extended with more articles and other QCL properties such as the layer sequences, thickness, the current density among others. There is also the need for more analysis of the proposed approach on other QCL properties with more diverse datasets. Future works may include extending the KG with more data, concepts and proposing learning methods for the QCL laser working properties prediction based on design features.
\section*{Availability of Materials\label{sec:availability}}
The source code and the materials used for the production of this work are publicly available at our GitHub repository: \url{https://github.com/DeperiasKerre/qKG}. 

\section*{Acknowledgement}
This work was funded by the French Embassy in Kenya (Scientific and Academic Cooperation Department) and the CNRS (under the framework “Dispositif de Soutien aux Collaborations avec l’Afrique sub-saharienne"). The authors would also like to thank Strathmore university, School of Computing and Engineering sciences and the Doctoral academy for creating an opportunity for this work to be produced.

\appendix
\section{Appendix: Sample Generated Prompt}
\begin{figure}
\centering
\includegraphics[width=1.\textwidth]{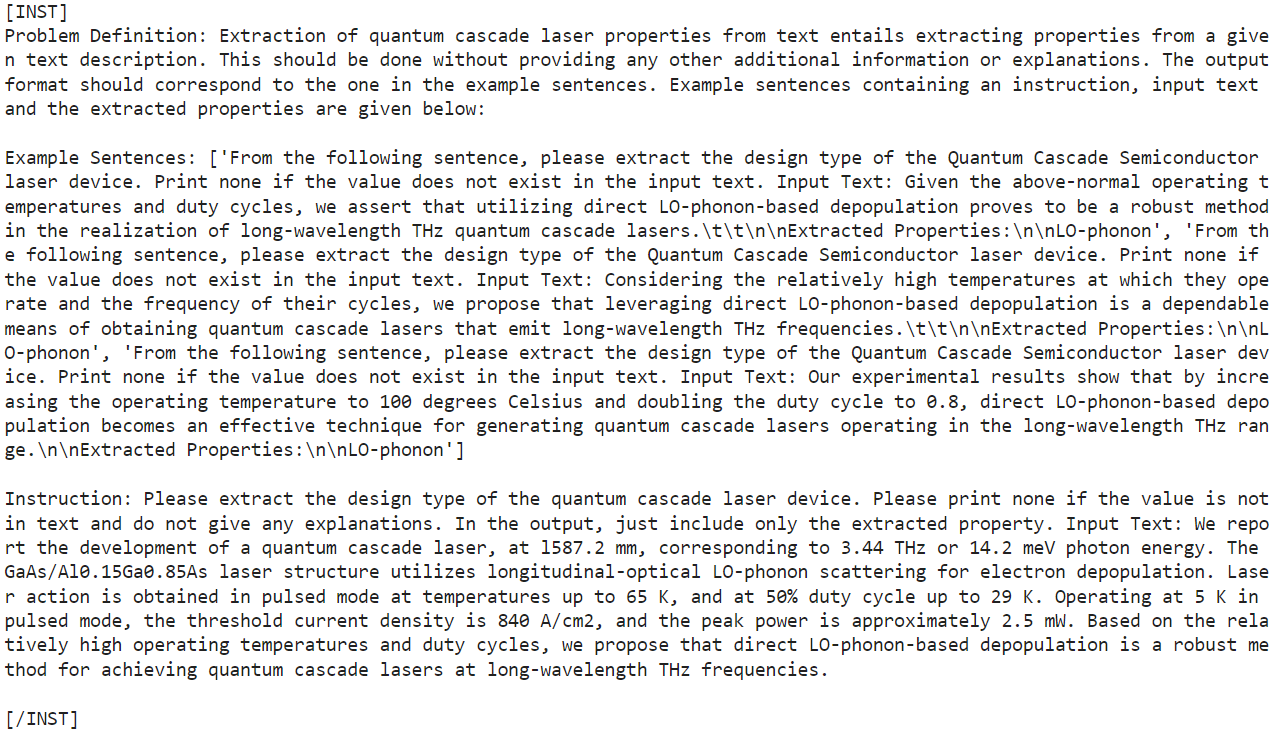}
\caption{Sample Regenerated Prompt} \label{fig:renegerated_prompt}
\end{figure}

\begin{thebibliography}{8}
\bibitem{ref_article1}
Faist, J., Capasso, F., Sivco, D. L., Hutchinson, A. L., Sirtori, C., \& Cho, A. Y. (1995). Quantum cascade laser: a new optical source in the mid-infrared. Infrared Physics \& Technology, 36(1), 99-103. \url{https://doi.org/10.1016/1350-4495(94)00058-S}
\bibitem{ref_article2}
Mittleman, D. M., Jacobsen, R. H., Neelamani, R., Baraniuk, R. G., \& Nuss, M. C. (1998). Gas sensing using terahertz time-domain spectroscopy. Applied Physics B: Lasers \& Optics, 67(3). \url{https://doi.org/10.1007/s003400050520} 
\bibitem{ref_article3}
Federici, J. F., Schulkin, B., Huang, F., Gary, D., Barat, R., Oliveira, F., \& Zimdars, D. (2005). THz imaging and sensing for security applications—explosives, weapons and drugs. Semiconductor science and technology, 20(7), S266. \url{https://doi.org/10.1088/0268-1242/20/7/018}
\bibitem{ref_article4}
Jepsen, P. U., Cooke, D. G., \& Koch, M. (2011). Terahertz spectroscopy and imaging–Modern techniques and applications. Laser \& Photonics Reviews, 5(1), 124-166. \url{ https://doi.org/10.1002/lpor.201000011}
\bibitem{ref_article5}
Wubs, J. R., Macherius, U., Weltmann, K. D., Lü, X., Röben, B., Biermann, K., ... \& Van Helden, J. H. (2023). Terahertz absorption spectroscopy for measuring atomic oxygen densities in plasmas. Plasma Sources Science and Technology, 32(2), 025006. \url{https://doi.org/10.1088/1361-6595/acb815}
\bibitem{ref_article6}
Richter, H., Wienold, M., Schrottke, L., Biermann, K., Grahn, H. T., \& Hübers, H. W. (2015). 4.7-THz local oscillator for the GREAT heterodyne spectrometer on SOFIA. IEEE Transactions on Terahertz Science and Technology, 5(4), 539-545. \url{https://doi.org/10.1109/TTHZ.2015.2442155}
\bibitem{ref_article7}
Shur, M., \& Liu, X. (2022, March). Biomedical applications of terahertz technology. In Advances in Terahertz Biomedical Imaging and Spectroscopy (Vol. 11975, p. 1197502). SPIE. \url{https://doi.org/10.1117/12.2604800}
\bibitem{ref_article8}
Ohtani, K., Turčinková, D., Bonzon, C., Benea-Chelmus, I. C., Beck, M., Faist, J., ... \& Stutzki, J. (2016). High performance 4.7 THz GaAs quantum cascade lasers based on four quantum wells. New Journal of Physics, 18(12), 123004. \url{https://doi.org/10.1088/1367-2630/18/12/123004}
\bibitem{ref_article9}
Kerre, D., Laurent, A., Maussang, K., \& Owuor, D. (2023, August). A text mining pipeline for mining the quantum cascade laser properties. In European Conference on Advances in Databases and Information Systems (pp. 393-406). Cham: Springer Nature Switzerland. \url{https://doi.org/10.1007/978-3-031-42941-5_34}
\bibitem{ref_article10}
Kerre, D., Laurent, A., Maussang, K., \& Owuor, D.(2024). A Concise Ontological Model of the Design and Optoelectronic Properties in the Quantum Cascade Laser Domain. (Preprint-Under Review).\url{http://dx.doi.org/10.13140/RG.2.2.36315.13608}
\bibitem{ref_article11}
Wilkinson, M. D., Dumontier, M., Aalbersberg, I. J., Appleton, G., Axton, M., Baak, A., ... \& Mons, B. (2016). The FAIR Guiding Principles for scientific data management and stewardship. Scientific data, 3(1), 1-9. \url{https://doi.org/10.1038/sdata.2016.18}
\bibitem{ref_article12}
Swain, M. C., Cole, J. M. (2016). ChemDataExtractor: a toolkit for automated extraction of chemical information from the scientific literature. Journal of chemical information and modeling, 56(10), 1894-1904.\url{https://doi.org/10.1021/acs.jcim.6b00207}
\bibitem{ref_article13}
Lowe, D. M., Sayle, R. A. (2015). LeadMine: a grammar and dictionary driven approach to entity recognition. Journal of cheminformatics, 7(1), 1-9.\url{https://doi.org/10.1186/1758-2946-7-S1-S5}
\bibitem{ref_article14}
Hawizy, L., Jessop, D. M., Adams, N., Murray-Rust, P. (2011). ChemicalTagger: A tool for semantic text-mining in chemistry. Journal of cheminformatics, 3, 1-13. \url{https://doi.org/10.1186/1758-2946-3-17}
\bibitem{ref_article15}
Leaman, R., Wei, C. H., Lu, Z. (2015). tmChem: a high performance approach for chemical named entity recognition and normalization. Journal of cheminformatics, 7(1), 1-10. \url{https://doi.org/10.1186/1758-2946-7-S1-S3}
\bibitem{ref_article16}
Rockt¨aschel, T., Weidlich, M., Leser, U. (2012). ChemSpot: a hybrid system for chemical named entity recognition. Bioinformatics, 28(12), 1633-1640. \url{https://doi.org/10.1093/bioinformatics/bts183}
\bibitem{ref_article17}
Sierepeklis, O., Cole, J. M. (2022). A thermoelectric materials database auto-generated from the scientific literature using ChemDataExtractor. Scientific Data, 9(1), 648. \url{https://doi.org/10.1038/s41597-022-01752-1}
\bibitem{ref_article18}
Dong, Q., Cole, J. M. (2022). Auto-generated database of semiconductor band
gaps using chemdataextractor. Scientific Data, 9(1), 193. \url{https://doi.org/10.1038/s41597-022-01294-6}
\bibitem{ref_article19}
Zhao, J., Cole, J. M. (2022). A database of refractive indices and dielectric constants auto-generated using chemdataextractor. Scientific data, 9(1), 192. \url{https://doi.org/10.1038/s41597-022-01295-5}
\bibitem{ref_article20}
Mavracic, J., Court, C. J., Isazawa, T., Elliott, S. R., Cole, J. M. (2021). ChemDataExtractor 2.0: Autopopulated ontologies for materials science. Journal of Chemical Information and Modeling, 61(9), 4280 4289. \url{https://doi.org/10.1021/acs.jcim.1c00446}
\bibitem{ref_article21}
Huang, S., Cole, J. M. (2020). A database of battery materials auto-generated using ChemDataExtractor. Scientific Data, 7(1), 260. \url{https://doi.org/10.1038/s41597-020-00602-2}
\bibitem{ref_article22}
Court, C. J., Cole, J. M. (2018). Auto-generated materials database of Curie and N´eel temperatures via semi-supervised relationship extraction. Scientific data, 5(1), 1-12. \url{https://doi.org/10.1038/sdata.2018.111}
\bibitem{ref_article23}
Cruse, K., Trewartha, A., Lee, S., Wang, Z., Huo, H., He, T., ... Ceder, G. (2022). Text-mined dataset of gold nanoparticle synthesis procedures, morphologies, and size entities. Scientific Data, 9(1), 234. \url{https://doi.org/10.1038/s41597-022-01321-6}
\bibitem{ref_article24}
Kononova, O., Huo, H., He, T., Rong, Z., Botari, T., Sun, W., ... Ceder, G. (2019). Textmined dataset of inorganic materials synthesis recipes. Scientific data, 6(1), 203. \url{https://doi.org/10.1038/s41597-019-0224-1}
\bibitem{ref_article25}
Korvigo, I., Holmatov, M., Zaikovskii, A., Skoblov, M. (2018). Putting hands to rest: efficient deep CNN-RNN architecture for chemical named entity recognition with no hand-crafted rules. Journal of cheminformatics, 10(1), 1-10. \url{https://doi.org/10.1186/s13321-018-0280-0}
\bibitem{ref_article26}
Zhao, J., Huang, S., Cole, J. M. (2023). OpticalBERT and OpticalTable-SQA: Text-and
Table-Based Language Models for the Optical-Materials Domain. Journal of Chemical Information
and Modeling. \url{https://doi.org/10.1021/acs.jcim.2c01259}
\bibitem{ref_article27}
Huang, S., Cole, J. M. (2022). BatteryBERT: A Pretrained Language Model for Battery Database Enhancement. Journal of Chemical Information and Modeling, 62(24), 6365-6377. \url{https://doi.org/10.1021/acs.jcim.2c00035}
\bibitem{ref_article28}
Thway, M., Low, A. K., Khetan, S., Dai, H., Recatala-Gomez, J., Chen, A. P., \& Hippalgaonkar, K. (2024). Harnessing GPT-3.5 for text parsing in solid-state synthesis–case study of ternary chalcogenides. Digital Discovery, 3(2), 328-336. \url{https://doi.org/10.1039/D3DD00202K}
\bibitem{ref_article29}
Choi, J., \& Lee, B. (2024). Accelerating materials language processing with large language models. Communications Materials, 5(1), 13. \url{https://doi.org/10.1038/s43246-024-00449-9}
\bibitem{ref_article30}
Polak, M. P., \& Morgan, D. (2024). Extracting accurate materials data from research papers with conversational language models and prompt engineering. Nature Communications, 15(1), 1569. \url{https://doi.org/10.1038/s41467-024-45914-8}
\bibitem{ref_article31}
Ye, Y., Ren, J., Wang, S., Wan, Y., Razzak, I., Xie, T., \& Zhang, W. (2024). Construction of Functional Materials Knowledge Graph in Multidisciplinary Materials Science via Large Language Model. arXiv preprint arXiv:2404.03080. \url{https://doi.org/10.48550/arXiv.2404.03080} 
\bibitem{ref_article32}
McCusker, J. P., Keshan, N., Rashid, S., Deagen, M., Brinson, C., \& McGuinness, D. L. (2020, November). Nanomine: A knowledge graph for nanocomposite materials science. In International Semantic Web Conference (pp. 144-159). Cham: Springer International Publishing. \url{https://doi.org/10.1007/978-3-030-62466-8_10}
\bibitem{ref_article33}
Mrdjenovich, D., Horton, M. K., Montoya, J. H., Legaspi, C. M., Dwaraknath, S., Tshitoyan, V., ... \& Persson, K. A. (2020). Propnet: a knowledge graph for materials science. Matter, 2(2), 464-480. \url{https://doi.org/10.1016/j.matt.2019.11.013}
\bibitem{ref_article34}
Nie, Z., Liu, Y., Yang, L., Li, S., \& Pan, F. (2021). Construction and application of materials knowledge graph based on author disambiguation: revisiting the evolution of LiFePO4. Advanced Energy Materials, 11(16), 2003580. \url{https://doi.org/10.1002/aenm.202003580}
\bibitem{ref_article35}
Venugopal, V., \& Olivetti, E. (2024). MatKG: An autonomously generated knowledge graph in Material Science. Scientific Data, 11(1), 217. \url{https://doi.org/10.1038/s41597-024-03039-z}
\bibitem{ref_article36}
Zhao, X., Greenberg, J., McClellan, S., Hu, Y. J., Lopez, S., Saikin, S. K., ... \& An, Y. (2021, December). Knowledge graph-empowered materials discovery. In 2021 IEEE International Conference on Big Data (Big Data) (pp. 4628-4632). IEEE. \url{https://doi.org/10.1109/BigData52589.2021.9671503}
\bibitem{ref_article37}
Zhang, Y., Chen, F., Liu, Z., Ju, Y., Cui, D., Zhu, J., ... \& Su, Y. (2024). A materials terminology knowledge graph automatically constructed from text corpus. Scientific Data, 11(1), 600. \url{https://doi.org/10.1038/s41597-024-03448-0}
\bibitem{ref_article38}
Statt, M. J., Rohr, B. A., Guevarra, D., Suram, S. K., \& Gregoire, J. M. (2023). The materials experiment knowledge graph. Digital Discovery, 2(4), 909-914. \url{https://doi.org/10.1039/D3DD00067B}
\bibitem{ref_article39}
OpenAI. 2023. New Models and Developer Products Announced at DevDay. \url{https://openai.com/index/new-models-and-developer-products-announced-at-devday/}. Accessed: 2024-09-28.
\bibitem{ref_article40}
Gao, Y., Xiong, Y., Gao, X., Jia, K., Pan, J., Bi, Y., ... \& Wang, H. (2023). Retrieval-augmented generation for large language models: A survey. arXiv preprint arXiv:2312.10997. \url{https://doi.org/10.48550/arXiv.2312.10997}
\bibitem{ref_article41}
Kerre, D., Laurent, A., Maussang, K., \& Owuor, D., (2024). An Instruction Dataset for Extracting Quantum Cascade Laser Properties from Scientific Text. Recherche Data Gouv: DOI: 10.57745/U3U7XR. \url{https://doi.org/10.57745/U3U7XR}
\bibitem{ref_article42}
Kerre, Deperias and Laurent, Anne and Maussang, Kenneth and Owuor, Dickson Odhiambo, An Instruction Dataset for Extracting Quantum Cascade Laser Properties from Scientific Text (October 29, 2024). Available at SSRN: \url{https://papers.ssrn.com/sol3/papers.cfm?abstract_id=5002930} 
\bibitem{ref_article43}
Song, K., Tan, X., Qin, T., Lu, J., \& Liu, T. Y. (2020). Mpnet: Masked and permuted pretraining for language understanding. Advances in neural information processing systems, 33,
16857-16867. \url{https://doi.org/10.5555/3495724.3497138}
\bibitem{ref_article44}
Kumar, S., Williams, B. S., Hu, Q., \& Reno, J. L. (2006). 1.9 THz quantum-cascade lasers with one-well injector. Applied Physics Letters, 88(12). \url{https://doi.org/10.1063/1.2189671}
\bibitem{ref_article45}
Khanal, S., Reno, J. L., \& Kumar, S. (2015). 2.1 THz quantum-cascade laser operating up to 144 K based on a scattering-assisted injection design. Optics express, 23(15), 19689-19697. \url{https://doi.org/10.1364/OE.23.019689}
\bibitem{ref_article46}
Williams, B. S., Callebaut, H., Kumar, S., Hu, Q., \& Reno, J. L. (2003). 3.4-THz quantum cascade laser based on longitudinal-optical-phonon scattering for depopulation. Applied Physics Letters, 82(7), 1015-1017. \url{https://doi.org/10.1063/1.1554479}
\bibitem{ref_article47}
Kumar, S., Hu, Q., \& Reno, J. L. (2009). 186 K operation of terahertz quantum-cascade lasers based on a diagonal design. Applied Physics Letters, 94(13).\url{https://doi.org/10.1063/1.3114418}
\bibitem{ref_article48}
 Razavipour, S. G., Dupont, E., Chan, C. W. I., Xu, C., Wasilewski, Z. R., Laframboise, S. R., ... \& Ban, D. (2014). A high carrier injection terahertz quantum cascade laser based on indirectly pumped scheme. Applied Physics Letters, 104(4). \url{https://doi.org/10.1063/1.4862177}
\bibitem{ref_article49}
Dupont, E., Fathololoumi, S., Wasilewski, Z. R., Aers, G., Laframboise, S. R., Lindskog, M., ... \& Liu, H. C. (2012). A phonon scattering assisted injection and extraction based terahertz quantum cascade laser. Journal of Applied Physics, 111(7).\url{https://doi.org/10.1063/1.3702571}
\bibitem{ref_article50}
Hempel, M., Röben, B., Niehle, M., Schrottke, L., Trampert, A., \& Grahn, H. T. (2017). Continuous tuning of two-section, single-mode terahertz quantum-cascade lasers by fiber-coupled, near-infrared illumination. AIP Advances, 7(5). \url{https://doi.org/10.1063/1.4983030}
\bibitem{ref_article51}
Wang, T., Liu, J. Q., Chen, J. Y., Liu, Y. H., Liu, F. Q., Wang, L. J., \& Wang, Z. G. (2013). Continuous-wave operation of terahertz quantum cascade lasers at 3.2 THz. Chinese Physics Letters, 30(6), 064201. \url{https://doi.org/10.1088/0256-307X/30/6/064201}
\bibitem{ref_article52}
Lü, X., Röben, B., Schrottke, L., Biermann, K., \& Grahn, H. T. (2021). Correlation between frequency and location on the wafer for terahertz quantum-cascade lasers. Semiconductor Science and Technology, 36(3), 035012. \url{https://doi.org/10.1088/1361-6641/abdd4b}
\bibitem{ref_article53}
Khabibullin, R. A., Shchavruk, N. V., Pavlov, A. Y., Klochkov, A. N., Glinskiy, I. A., Tomosh, K. N., ... \& Zhukov, A. E. (2019). Design and fabrication of terahertz quantum cascade laser with double metal waveguide based on multilayer GaAs/AlGaAs heterostructures. In IOP Conference Series: Materials Science and Engineering (Vol. 475, No. 1, p. 012020). IOP Publishing.\url{https://doi.org/10.1088/1757-899X/475/1/012020}
\bibitem{ref_article54}
Wen, B., Xu, C., Wang, S., Wang, K., Tam, M. C., Wasilewski, Z., \& Ban, D. (2018). Dual-lasing channel quantum cascade laser based on scattering-assisted injection design. Optics express, 26(7), 9194-9204. \url{https://doi.org/10.1364/OE.26.009194}
\bibitem{ref_article55}
Ushakov, D. V., Afonenko, A. A., Afonenko, A. A., Khabibullin, R. A., Fadeev, M. A., Gavrilenko, V. I., \& Dubinov, A. A. (2024). Feasibility of GaAs/AlGaAs quantum cascade laser operating above 6 THz. Journal of Applied Physics, 135(13).
\url{https://doi.org/10.1063/5.0198236}
\bibitem{ref_article56}
Deutsch, C., Krall, M., Brandstetter, M., Detz, H., Andrews, A. M., Klang, P., ... \& Unterrainer, K. (2012). High performance InGaAs/GaAsSb terahertz quantum cascade lasers operating up to 142 K. Applied Physics Letters, 101(21). \url{https://doi.org/10.1063/1.4766915}
\bibitem{ref_article57}
Brandstetter, M., Deutsch, C., Krall, M., Detz, H., MacFarland, D. C., Zederbauer, T., ... \& Unterrainer, K. (2013). High power terahertz quantum cascade lasers with symmetric wafer bonded active regions. Applied Physics Letters, 103(17).\url{https://doi.org/10.1063/1.4826943}
\bibitem{ref_article58}
Li, Y. Y., Liu, J. Q., Liu, F. Q., Zhang, J. C., Zhai, S. Q., Zhuo, N., ... \& Wang, Z. G. (2016). High power-efficiency terahertz quantum cascade laser. Chinese Physics B, 25(8), 084206. \url{https://doi.org/10.1088/1674-1056/25/8/084206}
\bibitem{ref_article59}
Wang, X., Shen, C., Jiang, T., Zhan, Z., Deng, Q., Li, W., ... \& Duan, S. (2016). High-power terahertz quantum cascade lasers with  $\sim 0.23 W$ in continuous wave mode. Aip Advances, 6(7).\url{https://doi.org/10.1063/1.4959195}
\bibitem{ref_article60}
Brandstetter, M., Kainz, M. A., Zederbauer, T., Krall, M., Schönhuber, S., Detz, H., ... \& Unterrainer, K. (2016). InAs based terahertz quantum cascade lasers. Applied Physics Letters, 108(1).\url{https://doi.org/10.1063/1.4939551}
\bibitem{ref_article61}
Valmorra, F., Scalari, G., Ohtani, K., Beck, M., \& Faist, J. (2015). InGaAs/AlInGaAs THz quantum cascade lasers operating up to 195 K in strong magnetic field. New Journal of Physics, 17(2), 023050.\url{https://doi.org/10.1088/1367-2630/17/2/023050}
\bibitem{ref_article62}
Walther, C., Scalari, G., Faist, J., Beere, H., \& Ritchie, D. (2006). Low frequency terahertz quantum cascade laser operating from 1.6 to1. 8THz. Applied Physics Letters, 89(23). \url{https://doi.org/10.1063/1.2404598}
\bibitem{ref_article63}
Williams, B. S., Kumar, S., Hu, Q., \& Reno, J. L. (2005). Operation of terahertz quantum-cascade lasers at 164 K in pulsed mode and at 117 K in continuous-wave mode. Optics express, 13(9), 3331-3339.\url{https://doi.org/10.1364/OPEX.13.003331}
\bibitem{ref_article64}
Walther, C., Fischer, M., Scalari, G., Terazzi, R., Hoyler, N., \& Faist, J. (2007). Quantum cascade lasers operating from 1.2 to1. 6THz. Applied Physics Letters, 91(13).\url{https://doi.org/10.1063/1.2793177}
\bibitem{ref_article65}
Hu, Q., Williams, B. S., Kumar, S., Callebaut, H., Kohen, S., \& Reno, J. L. (2005). Resonant-phonon-assisted THz quantum-cascade lasers with metal–metal waveguides. Semiconductor science and technology, 20(7), S228.\url{https://doi.org/10.1088/0268-1242/20/7/013}
\bibitem{ref_article66}
Olariu, T., Senica, U., \& Faist, J. (2024). Single-mode, surface-emitting quantum cascade laser at $26 \mu$m. Applied Physics Letters, 124(4).\url{https://doi.org/10.1063/5.0176281}
\bibitem{ref_article67}
Scalari, G., Amanti, M. I., Fischer, M., Terazzi, R., Walther, C., Beck, M., \& Faist, J. (2009). Step well quantum cascade laser emitting at 3 THz. Applied physics letters, 94(4).\url{https://doi.org/10.1063/1.3068496}
\bibitem{ref_article68}
Lever, L., Hinchcliffe, N. M., Khanna, S. P., Dean, P., Ikonić, Z., Evans, C. A., ... \& Kelsall, R. W. (2009). Terahertz ambipolar dual-wavelength quantum cascade laser. Optics express, 17(22), 19926-19932.\url{https://doi.org/10.1364/OE.17.019926}
\bibitem{ref_article69}
Schrottke, L., Lü, X., Rozas, G., Biermann, K., \& Grahn, H. T. (2016). Terahertz GaAs/AlAs quantum-cascade lasers. Applied Physics Letters, 108(10).\url{https://doi.org/10.1063/1.4943657}
\bibitem{ref_article70}
Fasching, G., Benz, A., Unterrainer, K., Zobl, R., Andrews, A. M., Roch, T., ... \& Strasser, G. (2005). Terahertz microcavity quantum-cascade lasers. Applied Physics Letters, 87(21).\url{https://doi.org/10.1063/1.2136222}
\bibitem{ref_article71}
Ohtani, K., Beck, M., Scalari, G., \& Faist, J. (2013). Terahertz quantum cascade lasers based on quaternary AlInGaAs barriers. Applied Physics Letters, 103(4).\url{https://doi.org/10.1063/1.4816352}
\bibitem{ref_article72}
Deutsch, C., Benz, A., Detz, H., Klang, P., Nobile, M., Andrews, A. M., ... \& Unterrainer, K. (2010). Terahertz quantum cascade lasers based on type II InGaAs/GaAsSb/InP. Applied Physics Letters, 97(26).\url{https://doi.org/10.1063/1.3532106}
\bibitem{ref_article73}
Li, L., Chen, L., Zhu, J., Freeman, J., Dean, P., Valavanis, A., ... \& Linfield, E. H. (2014). Terahertz quantum cascade lasers with >1 W output powers. Electronics letters, 50(4), 309-311.\url{https://doi.org/10.1049/el.2013.4035}
\bibitem{ref_article74}
Belkin, M. A., Fan, J. A., Hormoz, S., Capasso, F., Khanna, S. P., Lachab, M., ... \& Linfield, E. H. (2008). Terahertz quantum cascade lasers with copper metal-metal waveguides operating up to 178 K. Optics express, 16(5), 3242-3248.\url{https://doi.org/10.1364/OE.16.003242}
\bibitem{ref_article75}
Williams, B. S., Kumar, S., Qin, Q., Hu, Q., \& Reno, J. L. (2006). Terahertz quantum cascade lasers with double-resonant-phonon depopulation. Applied physics letters, 88(26).\url{https://doi.org/10.1063/1.2216112}
\bibitem{ref_article76}
Adams, R. W., Vijayraghavan, K., Wang, Q. J., Fan, J., Capasso, F., Khanna, S. P., ... \& Belkin, M. A. (2010). GaAs/Al0. 15Ga0. 85As terahertz quantum cascade lasers with double-phonon resonant depopulation operating up to 172 K. Applied Physics Letters, 97(13).\url{https://doi.org/10.1063/1.3496035}
\bibitem{ref_article77}
Williams, B. S., Kumar, S., Callebaut, H., Hu, Q., \& Reno, J. L. (2003). Terahertz quantum-cascade laser at $\lambda \approx 100 \micro $m using metal waveguide for mode confinement. Applied Physics Letters, 83(11), 2124-2126.\url{https://doi.org/10.1063/1.1611642}
\bibitem{ref_article78}
Luo, H., Laframboise, S. R., Wasilewski, Z. R., Aers, G. C., Liu, H. C., \& Cao, J. C. (2007). Terahertz quantum-cascade lasers based on a three-well active module. Applied physics letters, 90(4). \url{https://doi.org/10.1063/1.2437071}
\bibitem{ref_article79}
Adams, R. W., Vizbaras, A., Jang, M., Grasse, C., Katz, S., Boehm, G., ... \& Belkin, M. A. (2011). Terahertz sources based on intracavity frequency mixing in mid-infrared quantum cascade lasers with passive nonlinear sections. Applied Physics Letters, 98(15).\url{https://doi.org/10.1063/1.3579260}
\bibitem{ref_article80}
Bosco, L., Franckié, M., Scalari, G., Beck, M., Wacker, A., \& Faist, J. (2019). Thermoelectrically cooled THz quantum cascade laser operating up to 210 K. Applied Physics Letters, 115(1).\url{https://doi.org/10.1063/1.5110305}
\bibitem{ref_article81}
Wu, Y., Shen, Y., Addamane, S., Reno, J. L., \& Williams, B. S. (2021). Tunable quantum-cascade VECSEL operating at 1.9 THz. Optics express, 29(21), 34695-34706.\url{https://doi.org/10.1364/OE.438636}
\bibitem{ref_article82}
Li, J., Wan, T., \& Chen, C. (2019). Two-phonon-resonance terahertz quantum cascade laser based on GaN/AlGaN material system. Semiconductor Science and Technology, 34(7), 075018.\url{https://doi.org/10.1088/1361-6641/ab1401}
\bibitem{ref_article83}
Lander Gower, N., Levy, S., Piperno, S., Addamane, S. J., Reno, J. L., \& Albo, A. (2023). Two-well injector direct-phonon terahertz quantum cascade lasers. Applied Physics Letters, 123(6). \url{https://doi.org/10.1063/5.0155250}
\bibitem{ref_article84}
Vitiello, M. S., Scamarcio, G., Spagnolo, V., Dhillon, S. S., \& Sirtori, C. (2007). Terahertz quantum cascade lasers with large wall-plug efficiency. Applied Physics Letters, 90(19).\url{https://doi.org/10.1063/1.2737129}
\bibitem{ref_article85}
Schilling-Wilhelmi, M., Ríos-García, M., Shabih, S., Gil, M. V., Miret, S., Koch, C. T., ... \& Jablonka, K. M. (2024). From Text to Insight: Large Language Models for Materials Science Data Extraction. arXiv preprint arXiv:2407.16867. \url{https://doi.org/10.48550/arXiv.2407.16867}
\bibitem{ref_article86}
Sirin, E., Parsia, B., Grau, B. C., Kalyanpur, A., \& Katz, Y. (2007). Pellet: A practical owl-dl reasoner. Journal of Web Semantics, 5(2), 51-53. \url{https://doi.org/10.1016/j.websem.2007.03.004}
\end{thebibliography}
\end{document}